\documentclass[a4paper, 10pt, conference]{ieeeconf}      
\IEEEoverridecommandlockouts                              
\usepackage{color}
\usepackage{comment}
\usepackage{graphics}
\usepackage{epsfig}
\usepackage{algorithmic}
\usepackage{algorithm}

\newcommand{\ve}[1]{\mbox{\boldmath $#1$}}

\title{\LARGE \bf
ICP-Based Pallet Tracking for Unloading on Inclined Surfaces by Autonomous Forklifts
}

\author{Takuro Kato$^{1}$ and Mitsuharu Morisawa$^{1}$%
\thanks{$^{1}$T.Kato and M.Morisawa are with TICO-AIST Cooperative Research Laboratory for Advanced Logistics, National Institute of Advanced Industrial Science and Technology (AIST), Tsukuba, Japan.
        {\tt\small takuro.katou@aist.go.jp}}%
}

\begin{document}
\maketitle
\thispagestyle{empty}
\pagestyle{empty}

\begin{abstract}
This paper proposes a control method for autonomous forklifts to unload pallets on inclined surfaces, enabling the fork to be withdrawn without dragging the pallets.
The proposed method applies the Iterative Closest Point (ICP) algorithm\cite{Rusinkiewicz2001} to point clouds measured from the upper region of the pallet and thereby tracks the relative position and attitude angle difference between the pallet and the fork during the unloading operation in real-time.
According to the tracking result, the fork is aligned parallel to the target surface. 
After the fork is aligned, it is possible to complete the unloading process by withdrawing the fork along the tilt, preventing any dragging of the pallet.
The effectiveness of the proposed method is verified through dynamic simulations and experiments using a real forklift that replicate unloading operations onto the inclined bed of a truck.
\end{abstract}

\section{Introduction}
Due to the rising demand for e-commerce and the decline in the labor force, logistics automation is increasingly important.
Research and development in the automated control of forklifts, an indispensable part of logistics operations, have been actively conducted in recent years.

In indoor environments such as logistics warehouses, it can be assumed that the angle of the racks where pallets are unloaded is horizontal relative to the forklift.
In such cases, a forklift completes unloading operations by lowering the pallet onto the location while keeping the fork horizontal, and then withdrawing the fork straight out horizontally.

However, in semi-outdoor or outdoor environments, forklifts often need to adapt to variations in the angle of the location.
For example, when multiple heavy items are loaded onto a wing truck for shipment from a logistics facility, the truck bed may become tilted due to the weight affecting the vehicle's suspension.
Similarly, in environments where the ground is sloped for water drainage, the forklift itself can become tilted.
In such environments, to unload pallets onto the target surface and withdraw the fork without dragging the pallets, an unloading control method that explicitly takes into account the attitude difference between the forklift and the target surface is needed.

\begin{figure}
  \centering
  \includegraphics[width=8cm]{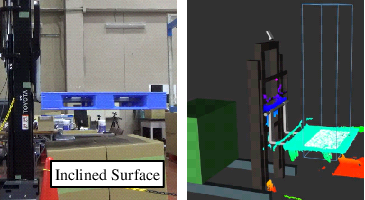}
  \caption{Autonomous unloading on inclined surface with our ICP based pallet tracking method. Right image indicates point cloud measured by RGBD camera.}
  \label{fig:fig1}
\end{figure}

This paper introduces a control method for unloading that aligns the tilt angle of the fork with that of inclined surfaces, illustrated in the left image of Fig. \ref{fig:fig1}.
The method applies the ICP algorithm\cite{Rusinkiewicz2001} on point clouds measured from the pallet's upper region to track the pallet in short time cycles, as depicted in the right image of Fig. \ref{fig:fig1}.
This method ensures the fork can be withdrawn without dragging the pallets, even on tilted surfaces.
The contributions of this paper to unloading control for autonomous forklifts are summarized as follows:
\begin{itemize}
\item A method for real-time estimation of attitude changes in pallets during unloading, achieved by executing the ICP algorithm at short time intervals.
\item A control algorithm for unloading pallets onto tilted surface, developed based on the aforementioned estimates.
\item Validation of the proposed method's effectiveness through dynamics simulations and experiments with a real forklift.
\end{itemize}

\section{Related Work}
Numerous studies have explored loading control in autonomous forklift operations.
In \cite{Seelinger2006} and \cite{ykita2022}, the position and yaw angle of the front surface of pallets are estimated using camera image processing.
In \cite{Wang2016}, a line-structured light sensor is used to guide a forklift autonomously to the front of the pallet.
If pallets are stacked horizontally, an autonomous forklift can load them by applying these methods.
In \cite{Iinuma2020}, a control method is proposed for the automatic insertion of forks into inclined pallets using LiDAR to estimate the pallet's pitch inclination.
In \cite{nkita2022}, the automatic insertion of forks into inclined pallets is achieved only using a monocular camera.

Studies including \cite{Varga2015}, \cite{Iinuma2021}, and \cite{Walter2010} address unloading control for autonomous forklifts.
In \cite{Varga2015}, the position of the target storage area for unloading is estimated using a stereo camera.
In \cite{Iinuma2021}, a control method is described for precisely stacking metal cage pallets with four legs, aligning the legs with the edges of previously placed pallets, using several RGBD cameras.
In \cite{Walter2010}, an unloading control method is proposed using LiDAR mounted horizontally and vertically on the forklift, estimating the position and yaw angle of the truck bed from measurements on each side of the forklift mast.
While these techniques enable autonomous forklifts to unload pallets onto horizontal surfaces, they do not take into account any inclinations.
Thus, they are not applicable to inclined storage areas.

\begin{figure}
  \centering
  \includegraphics[width=8cm]{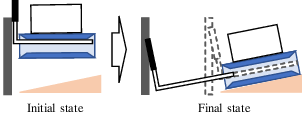}
  \caption{Initial and final states of unloading task}
  \label{fig:task_overview}
\end{figure}

Extending these methods to estimate the inclination and adjust the tilt angle of the fork could be an approach.
However, there are still challenges in adapting to changes in inclination during unloading due to the weight of the pallet.
This research focuses on the unique task of autonomous forklifts unloading pallets in front of an inclined storage area without dragging them. Fig. \ref{fig:task_overview} illustrates this task's initial and final states, clearly visualizing the problem.
Unlike previous studies, this proposal explicitly addresses the complexities of unloading pallets onto inclined surfaces.

\section{Pallet Unloading Control}
Generally, there is a gap between the pallet hole and the fork.
This gap causes the pallet to tilt relative to the fork when the pallet is unloaded on an inclined floor.
When the fork is moved so that they are parallel to the pallet, the pallet will follow a floor surface of unknown inclination.
Therefore, we design a pallet unloading controller in that the fork follows the tilt of the pallet by real-time tracking using ICP during the descent of the fork.

The proposed pallet unloading control is performed in the following steps.
\begin{enumerate}
\item Down the fork and put the pallet on the target place.
\item Adjust the fork tilt so that the fork can be aligned parallel to the target surface.
\item Pull the fork from the pallet along the tilt.
\end{enumerate}

We assume that the forklift is positioned in front of the target place before lowering a pallet.
The slope of the target place is unknown in advance or may be changed during unloading.

\subsection{Coordinate system}
In this paper, we suppose an RGBD camera is mounted on the top of the inner mast for the point cloud (PCL) measurement around the pallet on the fork. 
Here, the moving mechanical components on the chassis of the forklift are reach, inner mast, fork height, and fork tilt, respectively.
Then the transformation matrix of the RGBD camera ${^o}\ve{T}_{RGBD}$ in the origin frame $\sum_o$ which is defined on the chassis of the forklift, can be calculated from
\begin{equation}
    {^o}\ve{T}_{RGBD} = {^o}\ve{T}_{rch} \cdot {^{rch}}\ve{T}_{inr} \cdot {^{inr}}\ve{T}_{RGBD},
\end{equation}
where
\begin{eqnarray}
    {^i}\ve{T}_j =
    \left[ \begin{array}{cc}
        {^i}\ve{R}_j & {^i}\ve{p}_j \\
        {\ve{0}} & 1
    \end{array} \right] \in \Re^{4\times4} \nonumber
\end{eqnarray}
is $4\times4$ transformation matrix with rotation matrix ${^i}\ve{R}_j \in \Re^{3\times3}$ and position vector ${^i}\ve{p}_j \in \Re^3$.
${^{o}}\ve{T}_{rch}$ and ${^{rch}}\ve{T}_{inr}$ consist of reach and inner mast displacements respectively.
$^{inr}\ve{T}_{RGBD}$ is a constant transformation matrix which convert from the RGBD camera frame to inner mast frame.

Though the displacement of inner mast can't be measured directly due to hardware specifications, 
the inner mast and fork height are mechanically linked.
The inner mast displacement ${^{rch}}\ve{p}_{inr}$ can be converted from the fork height ${^{rch}}\ve{p}_{hgt}$ as follows. 
\begin{equation}
    {^{rch}}\ve{p}_{inr} = f({^{rch}}\ve{p}_{hgt}) 
\end{equation}
This relationship can be represented by two polygonal lines.
The inner mast moves at half the value of the fork in most ranges.

Similarly, the transformation matrix of the palette on the fork at the origin coordinate frame is obtained from
\begin{equation}
    {^o}\ve{T}_{pallet} = {^o}\ve{T}_{rch} \cdot {^{rch}}\ve{T}_{hgt} \cdot {^{hgt}}\ve{T}_{tilt} \cdot {^{tilt}}\ve{T}_{pallet},
\end{equation}
where ${^{hgt}}\ve{T}_{tilt}$ is calculated from fork tilt angle.
${^{tilt}}\ve{T}_{pallet}$ denotes the gazing point of the RGBD camera in the fork coordinate frame as a constant.

\subsection{Pallet tracking using ICP}
At the beginning of the fork down at a time $t_0$, the measured PCL $\ve{\chi}^{src}_{\cal{C}}(t_0)$ as the source PCL is extracted by bounding box (BB) in the camera coordinate frame $\sum_{\cal{C}}$.
The BB which is allocated in the area of the pallet size at a position of the fork, is projected to the camera frame.

Since it is unknown when the pallet will contact the floor, the pallet tracking for estimating the tilt angle and the height by ICP should be performed at short and constant cycles.
To reduce computational costs and to keep constant processing time, the source PCL is decimated to an almost constant number of points by random down sampling ($\approx 7000$ in this paper).

During the fork down, the convergence time of ICP also increases as the larger distance between the initial and the measured PCLs.
To solve this issue, the source PCL is translated in the camera coordinate frame by the amount of fork descent to make a new source PCL so that the it is closer to the actual position as
\begin{equation}
    {^{\cal{C}}}\bar{\ve{\chi}}^{src}(t) = \ve{\cal{P}}\left(^{C}\ve{T}_{hgt}(t)\right) {^{\cal{C}}}\ve{\chi}^{src}(t_0),
\end{equation}
where $\ve{\cal{P}}$ is transformation function that translates each PCL along with the fork descending.
Then we get the transformation matrix ${^C}\ve{T}^{icp}(t)$ to minimize error between the source PCL and the measured PCL by ICP as expressed as follows.
\begin{equation}
    {^C}\ve{T}^{icp}(t) = \arg\min \left(Error\left({^{\cal{C}}}\bar{\ve{\chi}}^{src}(t), {^{\cal{C}}}\ve{\chi}^{mes}(t) \right) \right)
\end{equation}
If the pallet doesn't contact on the floor, this matrix will be kept as the identity.
The difference height and tilt of the pallet can be extracted from the obtained ${^C}\ve{T}^{icp}(t)$ transformed to the origin frame and calculated as
\begin{equation}
    \left[ \begin{array}{cc}
        \Delta {^o\ve{R}}_{pallet}(t) & \Delta {^o\ve{p}}_{hgt}(t) \\
        {\ve{0}} & 1
    \end{array} \right] = {^o\ve{T}_{C}} \cdot {^{C}\ve{T}^{icp}(t)}.
\end{equation}
We define that the height of the fork moves along the Z axis and tilt rotates in the Y axis.
Each value ($\Delta tilt$,$\Delta height$) can be obtained as follows.
\begin{eqnarray}
    &&\Delta tilt(t) = \tan^{-1} \left( \frac{\ve{e}_x^{T} \cdot \Delta {^o\ve{R}}_{pallet}(t) \cdot \ve{e}_z}{\ve{e}_z^{T} \cdot \Delta {^o\ve{R}}_{pallet}(t) \cdot \ve{e}_z} \right) \\
    &&\Delta height(t) = \ve{e}_z^{T} \cdot \Delta {^o\ve{p}}_{hgt}(t)
\end{eqnarray}
Where $\ve{e}_{x}$ and $\ve{e}_{z}$ are the unit vector in axis of vertical and forward direction respectively.
By setting these differences to 0 with respect to the tilt angle, the fork will become parallel to the pallet.

\begin{figure}[bt]
  \centering
  \includegraphics[width=5cm]{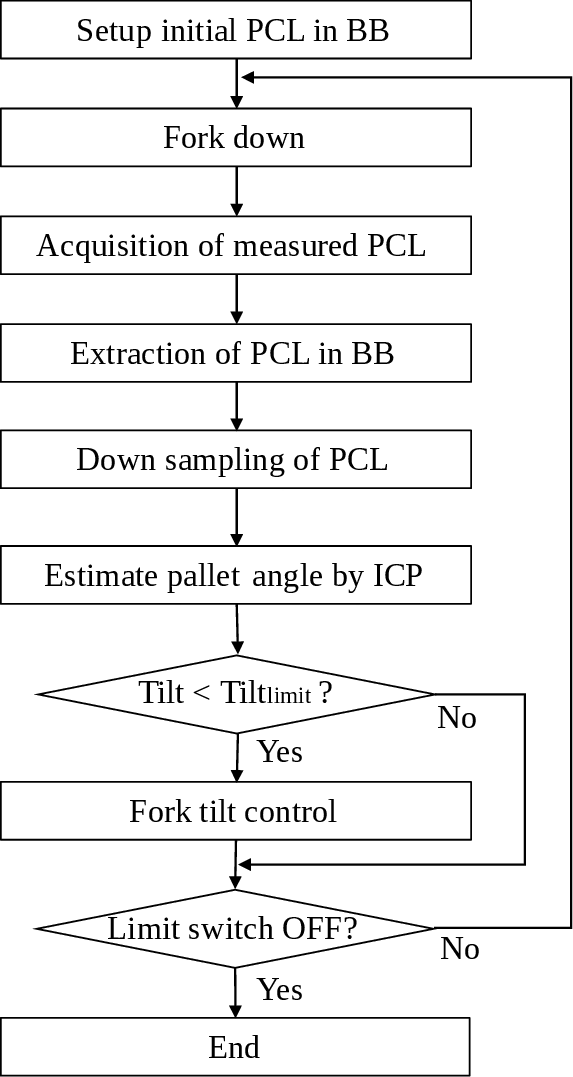}  
  \caption{Flowchart of pallet tracking control}
  \label{fig:pallet_tracking_control}
\end{figure}
First, a flowchart of the proposed pallet tracking control is shown in Fig. \ref{fig:pallet_tracking_control}.
Because our target forklift is driven by hydraulic actuators, it is difficult to control the position of the fork precisely.
So, in order to align the fork, the reference fork tile angle is updated from the current measured tilt angle when the relative angle between the pallet and fork estimated by ICP is above a threshold.

\begin{figure}[t]
  \centering
  \includegraphics[width=8cm]{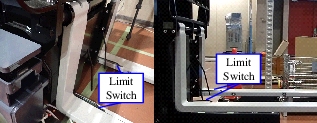}  
  \caption{Limit switches for loading detection attached with fork}
  \label{fig:limit_switch}
\end{figure}
If the fork and pallet are parallel, the fork can be withdrawn from the pallet if the fork is not in contact with the pallet.
Our forklift is equipped with a limit switch at the base of the fork that can detect when a pallet is loaded on the fork, as shown in Fig.\ref{fig:limit_switch}.
Although the height difference from ICP can be used to detect when the fork is off the pallet, we use this limit switch to detect unloading of the pallet.
The fork is then lowered until the limit switch is turned off.

\subsection{Fork withdraw control}
Once the fork is aligned parallel to the target surface, and the pallet is successfully placed there using the previously described method, the next step in the unloading operation is to withdraw the fork along its extension line in a straight and diagonal manner, while maintaining parallel alignment with the target surface.
The target trajectory for this withdrawal is determined using the fork's height and tilt angle sensor values at that moment.
The forklift is then moved backward at a constant speed.
The withdrawal amount of the fork is calculated from the forklift's wheel odometry, and the height error between the target trajectory and the fork is calculated using this withdrawal amount. Proportional control is applied to eliminate this height error, allowing the fork to follow the trajectory.
This ensures that the fork is withdrawn without dragging the pallet.

\section{Experiments}
Since experiments using a real forklift require a lot of development effort, a system that allows verification of algorithms in both a dynamics simulator and a real forklift will be constructed.
The simulations were conducted using Choreonoid\cite{Nakaoka2012}, an open-source dynamics simulator.
For the real-world experiments, a Toyota AGF Rinova 8AFBR15 was used.

\begin{figure}[htb]
  \centering
  \includegraphics[width=8cm]{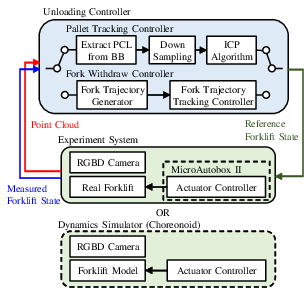}
  \caption{Overview of proposed system architecture}
  \label{fig:implementation_overview}
\end{figure}

\subsection{Experimental platform}
\begin{figure}[thb]
  \centering
  \includegraphics[width=7cm]{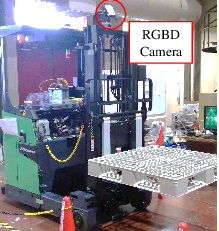}
  \caption{Experimental platform using Toyota Rinova AGF 8AFBR15 where the RGBD camera is mounted on the top of the mast.}
  \label{fig:platform}
\end{figure}

Fig. \ref{fig:implementation_overview} provides an overview of the experimental platform's implementation.
The system was designed to facilitate the use of a common controller for the dynamics simulator and the real forklift.
The pallet tracking controller was implemented as a nodelet in ROS as a common component between simulation and experiments.
An RGBD camera mounted on top of the forklift mast was utilized to measure the PCL, as illustrated in Fig. \ref{fig:platform}.

In the real-world experiments, we employed the Microsoft Azure Kinect as the RGBD camera, with the mode set to NFOV Unbinned (Resolution : 640$\times$576 px, FoI : 75$\times$65 deg).
The ICP algorithm was executed on an Nvidia Jetson Xavier, using the cuPCL library \cite{Nvidia2023} for the computation.
Additionally, we mounted the dSPACE MicroAutobox II (MAB) on the forklift as the controller and modified the forklift to enable control over the fork's position and angle based on control signals from the MAB.

The Azure Kinect was connected to the Jetson.
The Jetson and MAB communicated with each other via UDP.
The forklift was modified to transmit measurements of the fork state to MAB.
Specifically, this included the height of the fork (distance from the ground), reach displacement (extension or retraction of the fork), and tilt angle.
The measured and desired forklift states on the Jetson are communicated with the MAB via UDP at a rate of 50Hz.
These acquired forklift states are sent to the pallet tracking controller as a ROS message in each case of the dynamics simulation and the real forklift.
The Pallet Tracking Controller is executed in synchronization with the received PCL from Azure Kinect at a rate of 5Hz.

\subsection{Simulation} \label{subsec:simulation}
In real forklifts, the fork height and tilt are driven by hydraulic cylinders.
Due to the difficulty in simulating hydraulic cylinders precisely in Choreonoid, a simplified model is applied where each moving component is driven by force or torque through the PD controller.
In addition, the fork height is driven through the metal chain, so it is not possible to generate a pushing force greater than the fork's weight.
The simulation reproduces this one-way behavior by setting a force limit in the pushing direction. 
Furthermore, there is a time delay of actuation when the command value is given to the actuator, and this behavior is approximated through a first-order low pass filter with a large time constant, thereby reflecting these effects.

The simulation conditions are shown in Table \ref{table:experiment_simulation}.
In Cases 1 and 2, the truck is already tilted 3 deg with three 500 kg loads on the truck bed by the suspensions, and is tilted up to 4 deg with another 500 kg load on it by unloading.
In Cases 3 and 4, a cage containing cabbage is placed on a surface inclined in the opposite direction to that in Cases 1 and 2.
Cases 1 and 3 are when our proposed method is applied, while Cases 2 and 4 are when the fork tilt angle is not controlled.

\begin{table}[htbp]
  \caption{Simulation Conditions}
  \label{table:experiment_simulation}
  \centering
  \begin{tabular}{c|ccc}
    \hline
    Case ID  & Load on Pallet & Target surface & Controller \\
    \hline \hline
    1 & Cartoon Box & Up \& Variable & Proposed\\
    2 & Cartoon Box & Up \& Variable & No Control\\
    3 & Cabbage in Cage & Down & Proposed\\
    4 & Cabbage in Cage & Down & No Control\\
    \hline
  \end{tabular}
\end{table}

\begin{figure}[htbp]
  \centering
  \includegraphics[width=7.2cm]{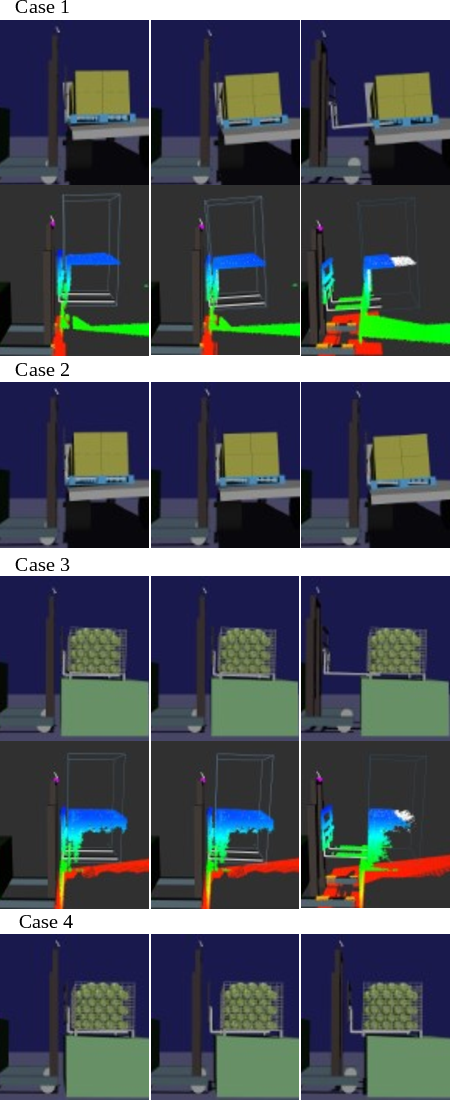}
  \caption{Simulation results}
  \label{fig:simulation}
\end{figure}

These simulations are shown in Fig. \ref{fig:simulation}.
In Cases 1 and 3, our controller adjusted the fork tilt angle adaptationally and completed the unloading task without dragging the pallet.
In Case 2, the fork remained in contact with the pallet, preventing the limit switch at the base of the fork from turning off.
As a result, the controller halted operations before proceeding to the fork withdrawal process.
In Case 4, the limit switch at the base of the fork turned off while the cage was still on the tip of the fork, and the withdrawal operation started.
Then, the dragging cage was observed.

\begin{figure}[t]
  \begin{tabular}{c}
    \begin{minipage}[t]{\linewidth}
      \centering
      \includegraphics[width = \columnwidth]{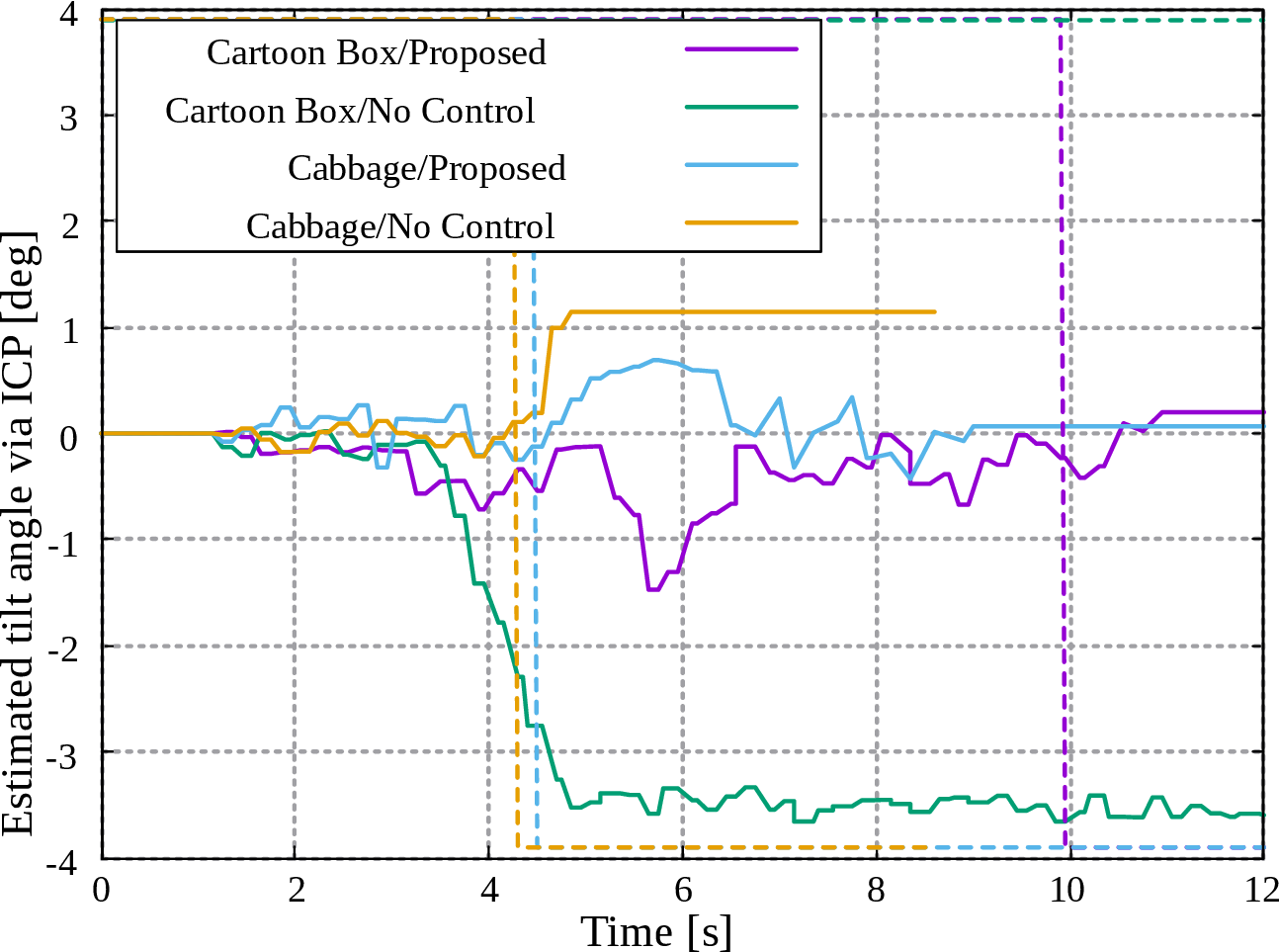}
      \caption{Estimated tilt angle via ICP}
      \label{fig:simulation_log_icp}
    \end{minipage} \\ \\
    \begin{minipage}[t]{\linewidth}
      \centering
      \includegraphics[width = \columnwidth]{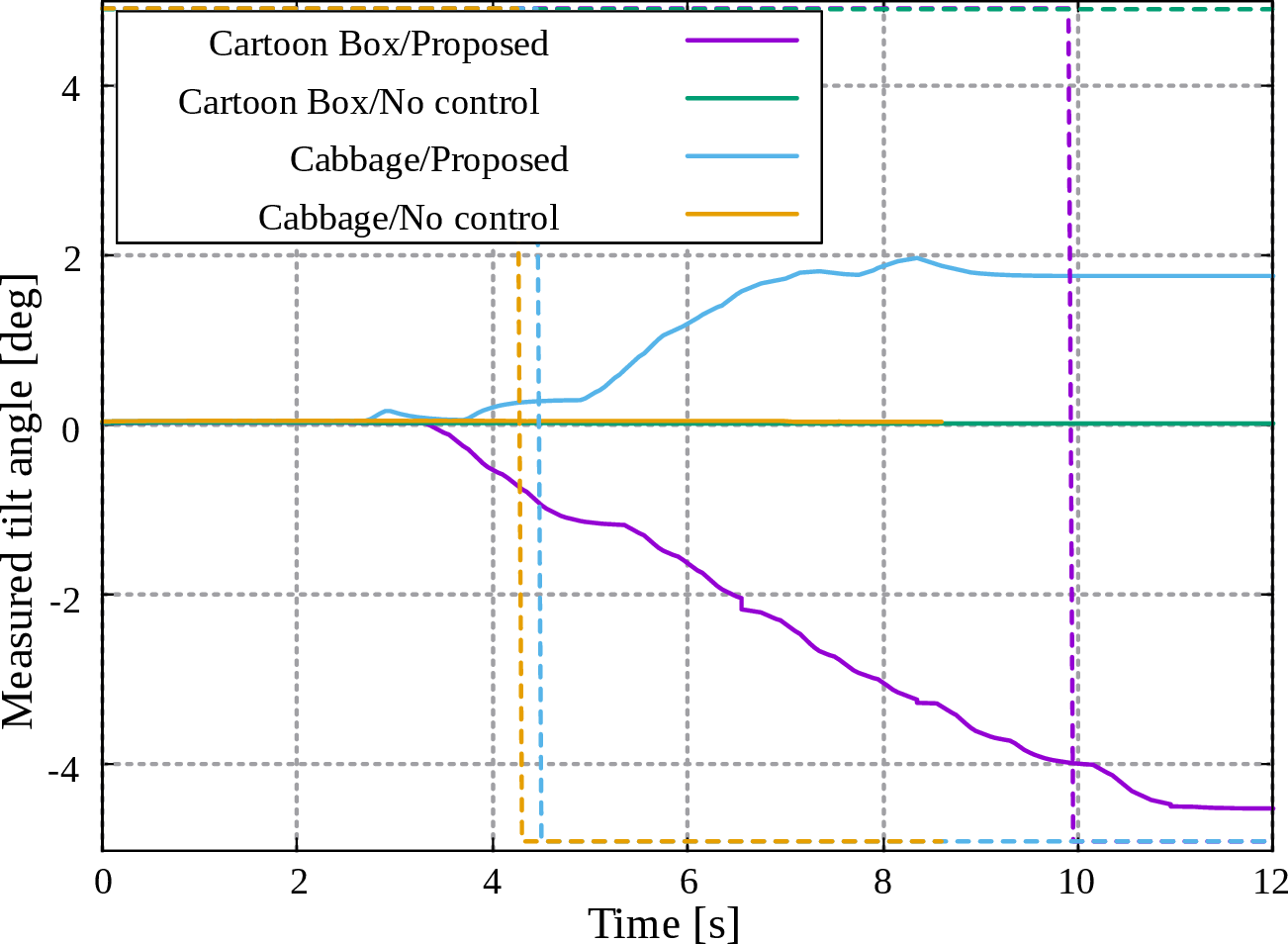}
      \caption{Measured tilt angle of fork}
      \label{fig:simulation_log_tilt}
    \end{minipage}
  \end{tabular}
\end{figure}

The difference between the tilt angles of the pallet and fork estimated by ICP is shown in Fig. \ref{fig:simulation_log_icp}.
The dashed line indicates the state of the limit switch, which is on (high) / off (low)  when a load / no load is detected on the fork.
Since ICP is not performed during the withdrawal operation, a constant value at the end time in the figure indicates that this operation is in progress.
In Cases 1 and 3, we can see that the difference between the pallet and fork converged below the threshold ($<$ 0.25 deg), and the withdrawal operation was started and completed.
In Case 2, where the pallet tracking control was not applied, the limit switch didn't turn off to start the withdrawal operation as described in Fig. \ref{fig:simulation}, and the difference remains at a certain value.
In Case 4, the pallet contacted the ground first on the near side of the forklift mast.
Then, the limit switch was turned off, and the withdrawal operation started with a constant difference in tilt angle between the fork and the pallet.

Fig. \ref{fig:simulation_log_tilt} shows the measured tilt angle of the fork that was moved.
In Cases 2 and 4, no control was applied to adjust the tilt angle; therefore, there is no change in the values.
In Cases 1 and 3, where the proposed pallet tracking control method was applied, the tilt angle of the fork was moved to around Case 1 : -4 deg, Case 3 : 2 deg, which was close to the final pallet tilt, and the withdrawing could be completed.

\subsection{Experiments with a real forklift}
The mock storage area was set in front of the forklift and was inclined at specific angles for the tests.
As detailed in Table \ref{table:experiment_actual_forklift}, experiments were performed under four distinct conditions.
In Cases 1 and 3, we set the tilt of the target surface to approximately 4 deg.
In Cases 2 and 4, we adjusted the inclination to about 2 deg, considering the forklift's tilt range.

\begin{table}[htb]
  \caption{Experimental Conditions}
  \label{table:experiment_actual_forklift}
  \centering
  \begin{tabular}{c|cc}
    \hline
     Case ID  & Load on Pallet & Target surface \\
    \hline
    \hline
    1 & Empty & Up\\
    2 & Empty & Down\\
    3 & Loaded & Up\\
    4 & Loaded & Down\\    
    \hline
  \end{tabular}
\end{table}

\begin{figure}[thp]
  \centering
  \includegraphics[width = 8cm]{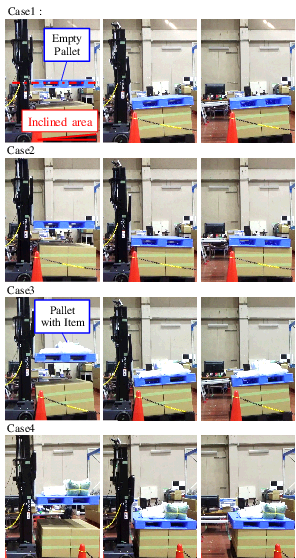}
  \caption{Snapshot of unloading operation}
  \label{fig:experiments_actual}
\end{figure}

Fig. \ref{fig:experiments_actual} shows the results.
The left image in each test case represents the state before the unloading operation begins.
The center image shows the state where the angle of the fork is aligned with the target surface, and the right image depicts the state after the unloading operation is complete.
In all cases, the forklift successfully completed the unloading operation without dragging the pallet.
The pallet tracking controller was executed at frequencies ranging from 2.5 to 5 Hz throughout the experiments.

\begin{figure}[tbp]
  \begin{tabular}{c}
    \begin{minipage}[t]{\linewidth}
      \centering
      \includegraphics[width = \columnwidth]{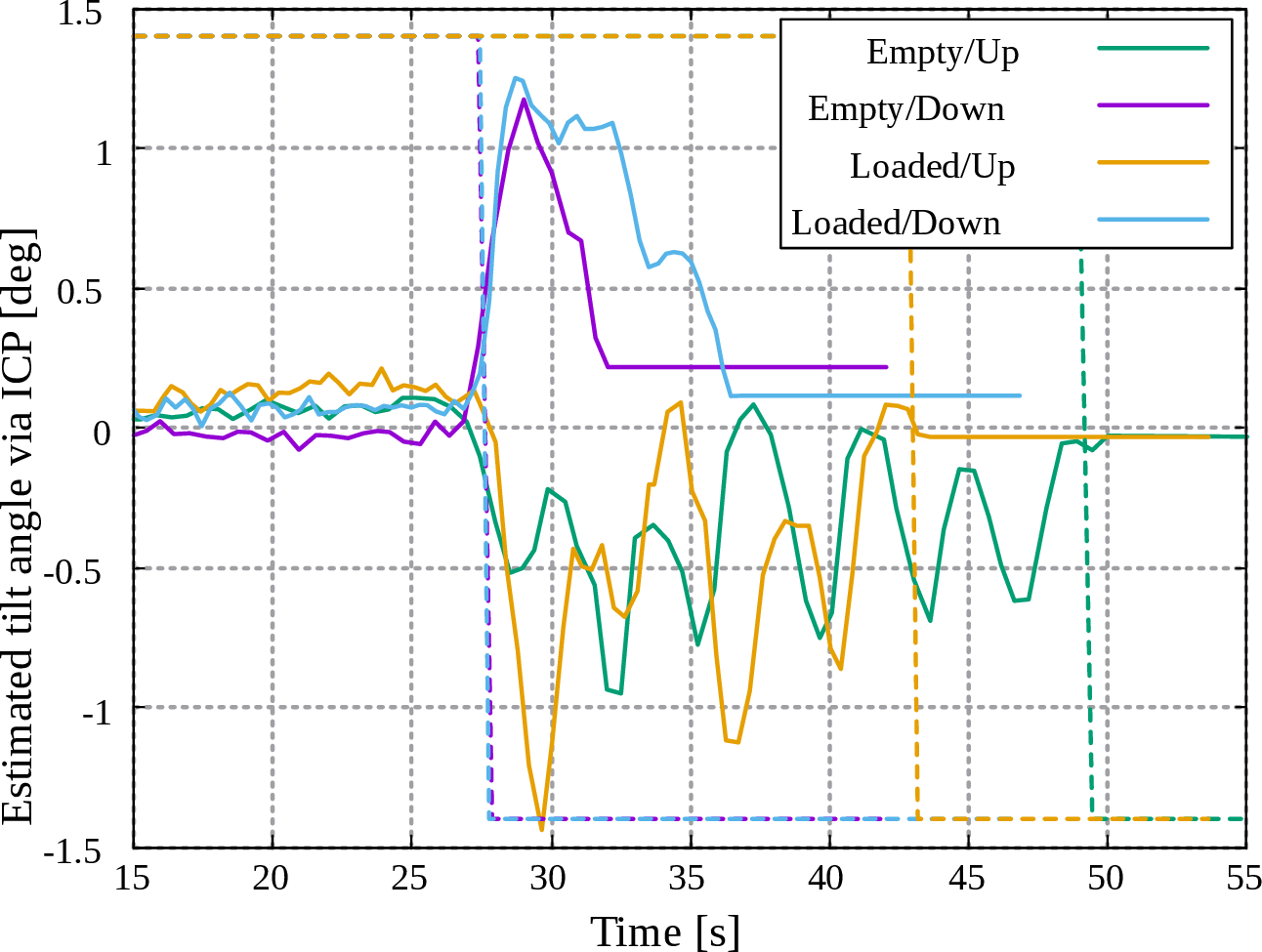}
      \caption{Estimated tilt angle via ICP}
      \label{fig:exmeriment_log_icp}
    \end{minipage} \\ \\
    \begin{minipage}[t]{\linewidth}
      \centering
      \includegraphics[width = \columnwidth]{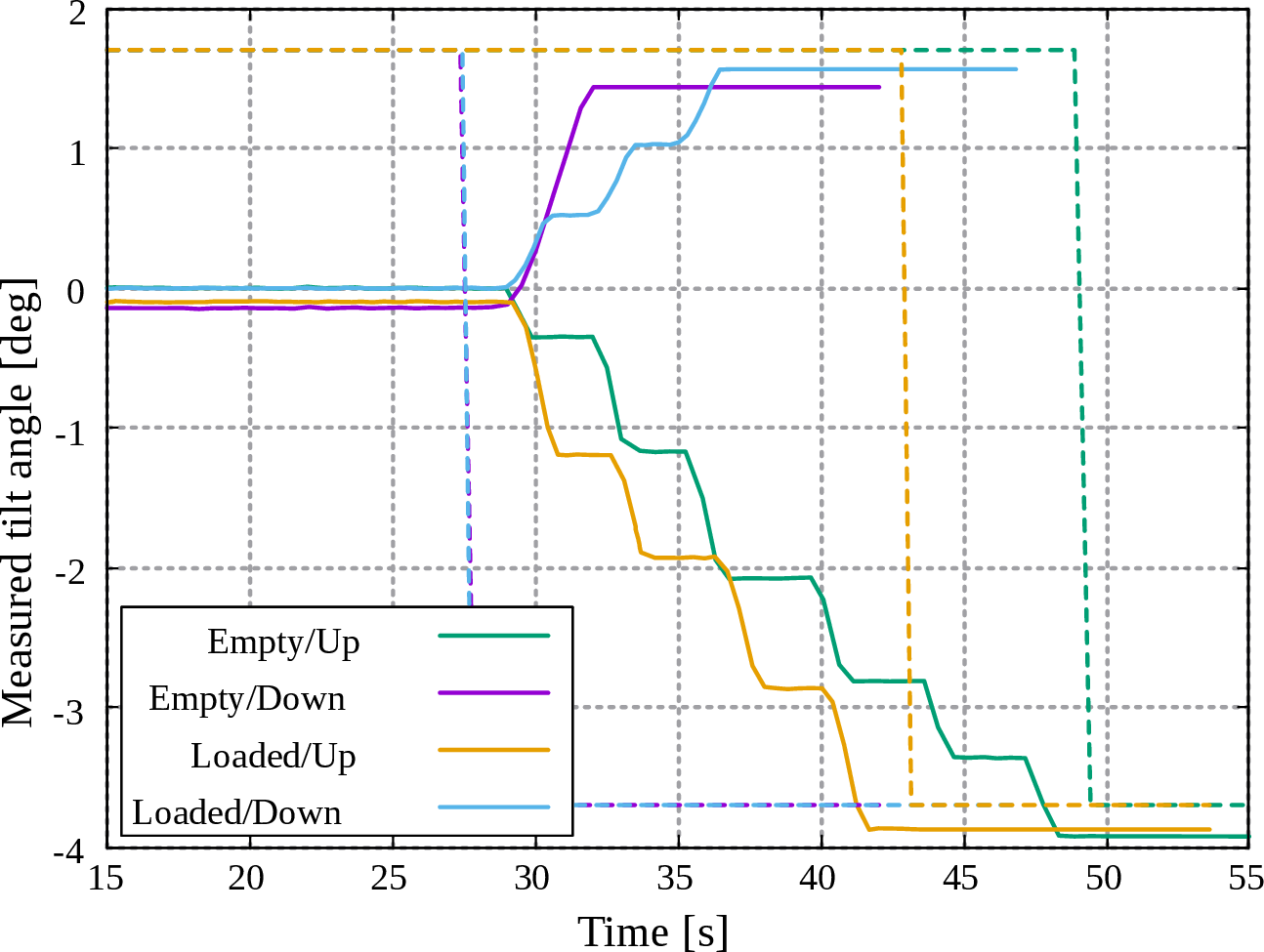}
      \caption{Measured tilt angle of fork}
      \label{fig:experiment_log_tilt}
    \end{minipage}
  \end{tabular}
\end{figure}

Fig. \ref{fig:exmeriment_log_icp} shows the estimated relative tilt angle between the tilt angles of the pallet and fork via ICP.
In all test cases, the estimations finally converged to within 0.25 deg.
We can see that the terminal tilt of the fork is close to the target surface in Fig. \ref{fig:experiment_log_tilt}.
The tilt angle adjustment and fork height adjustment are implemented to be performed alternately in this experiment; therefore, the measured tilt is staircase-like.
While the actual difference in angles between the pallet and the target surface will have the ICP estimation error added to it, we have confirmed through prior experiments in advance that by properly setting the BB region, ICP can be performed with sufficient accuracy.

The dashed line is the state of the limit switch as in the simulation in Sec. \ref{subsec:simulation}.
As well as in the simulation, in the case of the up inclination (Cases 1 and 3), we can see that the tilt angle of the fork is adjusted first.
Then, the withdrawal operation is started immediately after the limit switch is turned off.
In contrast, in the case of the down inclination (Cases 2 and 4), the fork tilt angle is adjusted after the limit switch is turned off first.
After the difference in tilt angle becomes smaller than the threshold, the withdrawal operation is started.
As a result, our controller successfully withdrew the fork from the pallet without dragging it in all cases.

\section{Conclusion}
This paper proposed a control method for autonomous forklifts to unload pallets onto inclined surfaces.
The proposed method uses the ICP algorithm to track the pallet and adjusts the tilt angle of the fork to eliminate the attitude angle difference between the fork and the pallet during the unloading operation.
By running the ICP algorithm in short time intervals, the method can track the pallet in real-time, enabling its application in environments where the tilt of the target surface changes during the unload operation.
Moreover, the proposed method is independent of the shapes of goods on pallets, accommodating even irregularly shaped items.
Simulations and real-world experiments validated the effectiveness of the proposed method.

The proposed method offers flexibility in selection and location to install the point cloud sensors.
Instead of using the RGBD cameras, LiDAR can be employed as the point cloud sensor.
The location to install the sensor need not be limited to the top of the forklift mast; it can also be installed on the forklift chassis.
In cases where it may be hard to measure a point cloud with sufficient shape features for ICP tracking using a single camera, such as when handling tall items on pallets, it is possible to combine multiple point cloud sensors.

The responsiveness of the proposed control method is dependent on the execution cycle of the ICP.
Therefore, it is expected that shortening the execution time reduces operational time.
Future work includes enhancing the operational speed and integrating our ongoing pallet loading techniques (i.e., automatic insertion of fork into inclined pallet\cite{nkita2022}) to advance the level of automation of forklift operation.

\end{document}